\pdfoutput=1

\documentclass[11pt]{article}

\usepackage[preprint]{acl}

\usepackage{times}
\usepackage{latexsym}
\usepackage{amssymb}
\usepackage[T1]{fontenc}
\usepackage{hyperref}

\usepackage[utf8]{inputenc}

\usepackage{microtype}

\usepackage{inconsolata}

\usepackage{graphicx}

\title{RoadscapesQA: A Multitask, Multimodal Dataset for Visual Question Answering on Indian Roads}

\author{
Vijayasri Iyer \\
\texttt{thisisvij98@gmail.com}
\And
Maahin Rathinagiriswaran \\
\texttt{rmaahin00@gmail.com}
\And
Jyothikamalesh S \\
\texttt{jyothikamaleshs@gmail.com}
}

\begin{document}
\maketitle
\begin{abstract}
Understanding road scenes is essential for autonomous driving, as it enables systems to interpret visual surroundings to aid in effective decision-making. We present Roadscapes, a multitask multimodal dataset consisting of upto 9,000 images captured in diverse Indian driving environments, accompanied by manually verified bounding boxes. To facilitate scalable scene understanding, we employ rule-based heuristics to infer various scene attributes, which are subsequently used to generate question-answer (QA) pairs for tasks such as object grounding, reasoning, and scene understanding. The dataset includes a variety of scenes from urban and rural India, encompassing highways, service roads, village paths, and congested city streets, captured in both daytime and nighttime settings. Roadscapes has been curated to advance research on visual scene understanding in unstructured environments. In this paper, we describe the data collection and annotation process, present key dataset statistics, and provide initial baselines for image QA tasks using vision-language models. Our dataset and code is an anonymized format available at: \href{https://github.com/vijpandaturtle/roadscapes} {https://github.com/vijpandaturtle/roadscapes}
\end{abstract}

\section{Introduction}
As sophisticated perception systems continue to advance, the development of computer vision and foundational models is increasingly oriented towards multi-task and multimodal architectures. These models integrate visual perception capabilities such as object detection and localization, semantic segmentation etc., with natural language understanding, enabling richer interactions between vision and language. This shift is largely driven by the promise of a deeper semantic understanding of complex driving environments. Autonomous driving has shifted to developing several vision-language systems for increased interpretability and to leverage the emergent abilities of foundation models. For this task, however, it requires access to high-fidelity data with a verifiable ground truth. The majority of driving multimodal datasets, containing vision and text annotation, cover regions like the United States, Europe and specific Asian countries like Singapore, China and Japan. These datasets, while comprehensive in some aspects, reflect road infrastructure, traffic behavior, and environmental conditions that differ significantly from those encountered in developing countries in the South Asian region. Roads in these countries are typically well-marked, consistently maintained, and governed by standardized traffic rules. In contrast, driving conditions in countries like India are far more variable, characterized by high traffic density, unpredictable agent behavior, heterogeneous vehicle types, unmarked roads, and frequent interactions with non-motorized agents such as pedestrians, cyclists, and animals.

\begin{figure}[t]
  \includegraphics[width=1.0\linewidth]{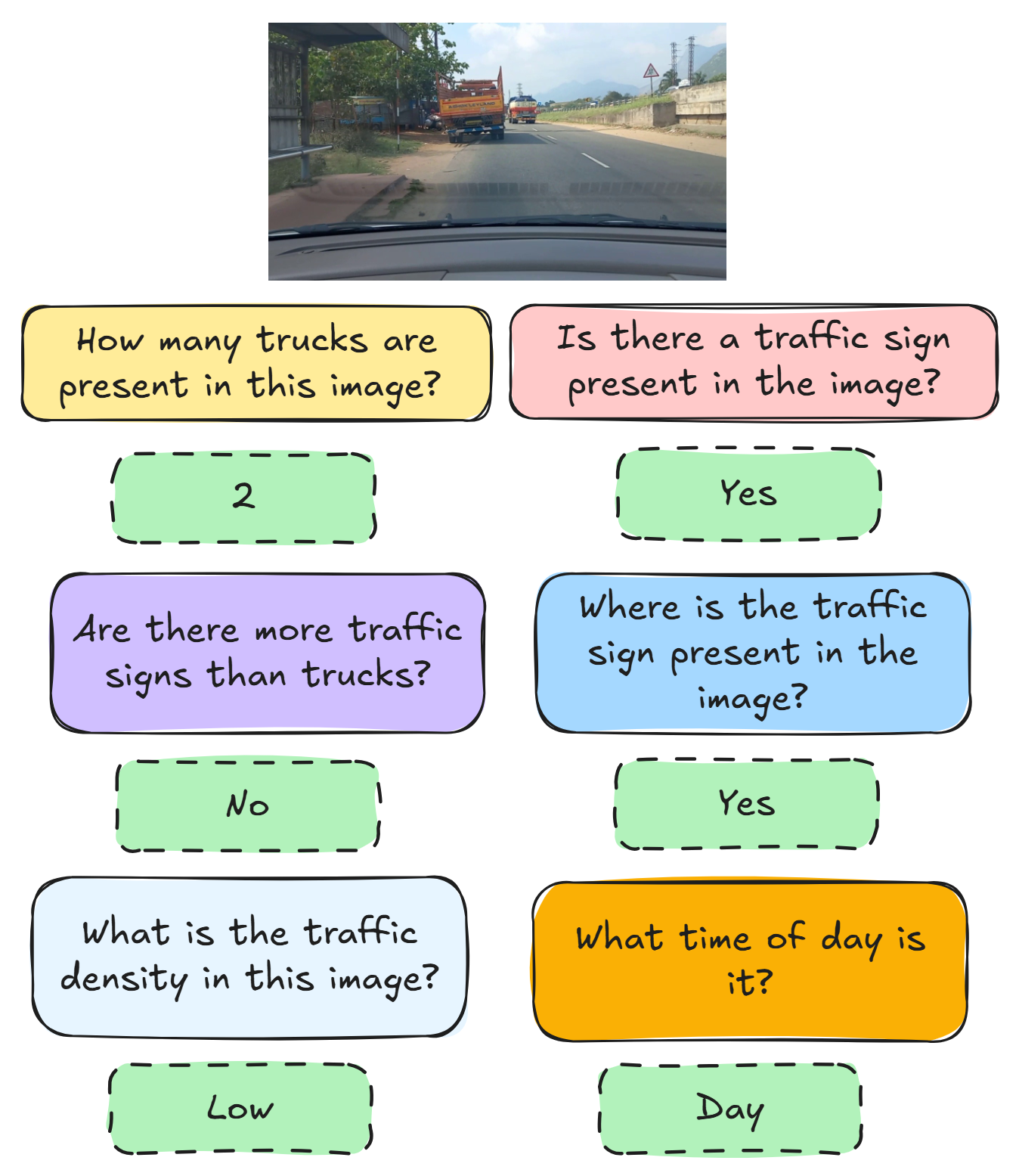} 
  \caption {A example of an image and corresponding questions from the VQA Dataset.}
\end{figure}

Construction of a dataset to improve diversity can take a significant amount of time and monetary investment (i.e an extremely high-cost sensor suite) to produce, which makes reproducibility or scaling these efforts very difficult. In this work, we aim to represent scenarios within Indian roads and highways that complement existing Indian datasets while filling the gap in presenting scenarios such as nighttime driving and rural environments. Instead of relying on an expensive sensor suit and several human annotators, we rely on a low-cost monocular camera, pre-annotation by state-of-the-art computer vision models for aiding in annotation, human verification followed by manually defined heuristics for scalable and automatic label generation. 

Roadscapes contains almost 9000 monocular images collected from a wide range of urban and rural regions in southern India, annotated for two computer vision tasks: object detection and road segmentation as well as Visual Question Answering (VQA) covering a variety of question related to scene understanding such as object counting, localization, object description, spatial relationship identification. The dataset captures a wide spectrum of driving environments, including highways, arterial roads, city streets, narrow rural paths, and mixed-use roadways. The dataset includes some sensor artifacts such as motion jitter, blur, glare, and shadowing effects caused by windshields and dashboard reflections. Such conditions are often ignored in high-end datasets captured using expensive, stabilized rigs, but are crucial for developing models that operate in resource-constrained settings. 

Our contributions are as follows. (1) A diverse image and video dataset consisting of over 9,000 images covering a wide range of conditions—including urban and rural environments across India, and varied lighting scenarios such as daytime and nighttime. (2) A generation framework for image-level question-answer pairs, combining computer vision annotations, heuristic rules, and scene graphs inferred from large language models. (3) A set of baselines evaluating image-level understanding via question-answering, grounded in real-world Indian driving footage.

\begin{figure*}[t]
  \includegraphics[width=0.48\linewidth]{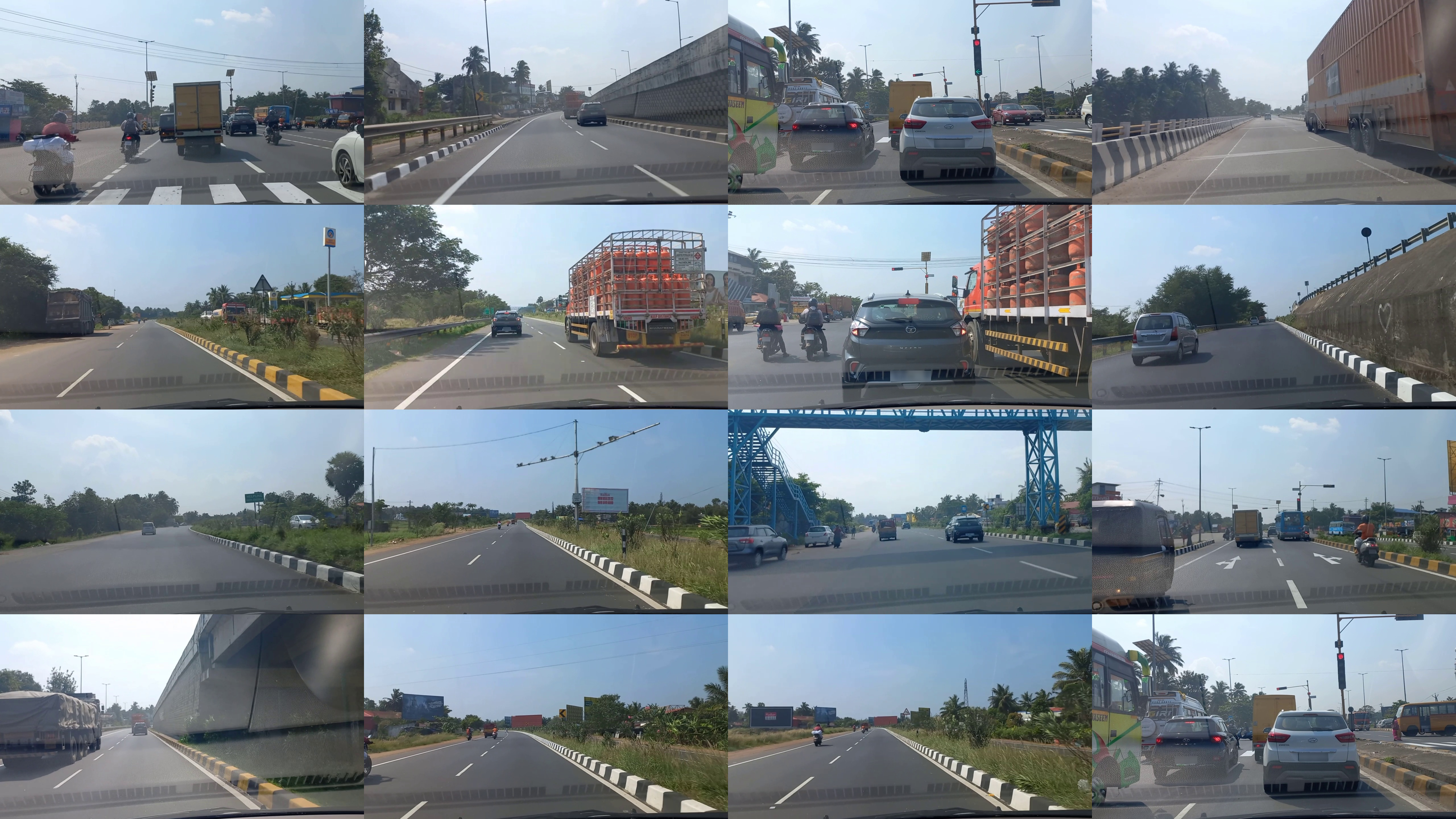} \hfill
  \includegraphics[width=0.48\linewidth]{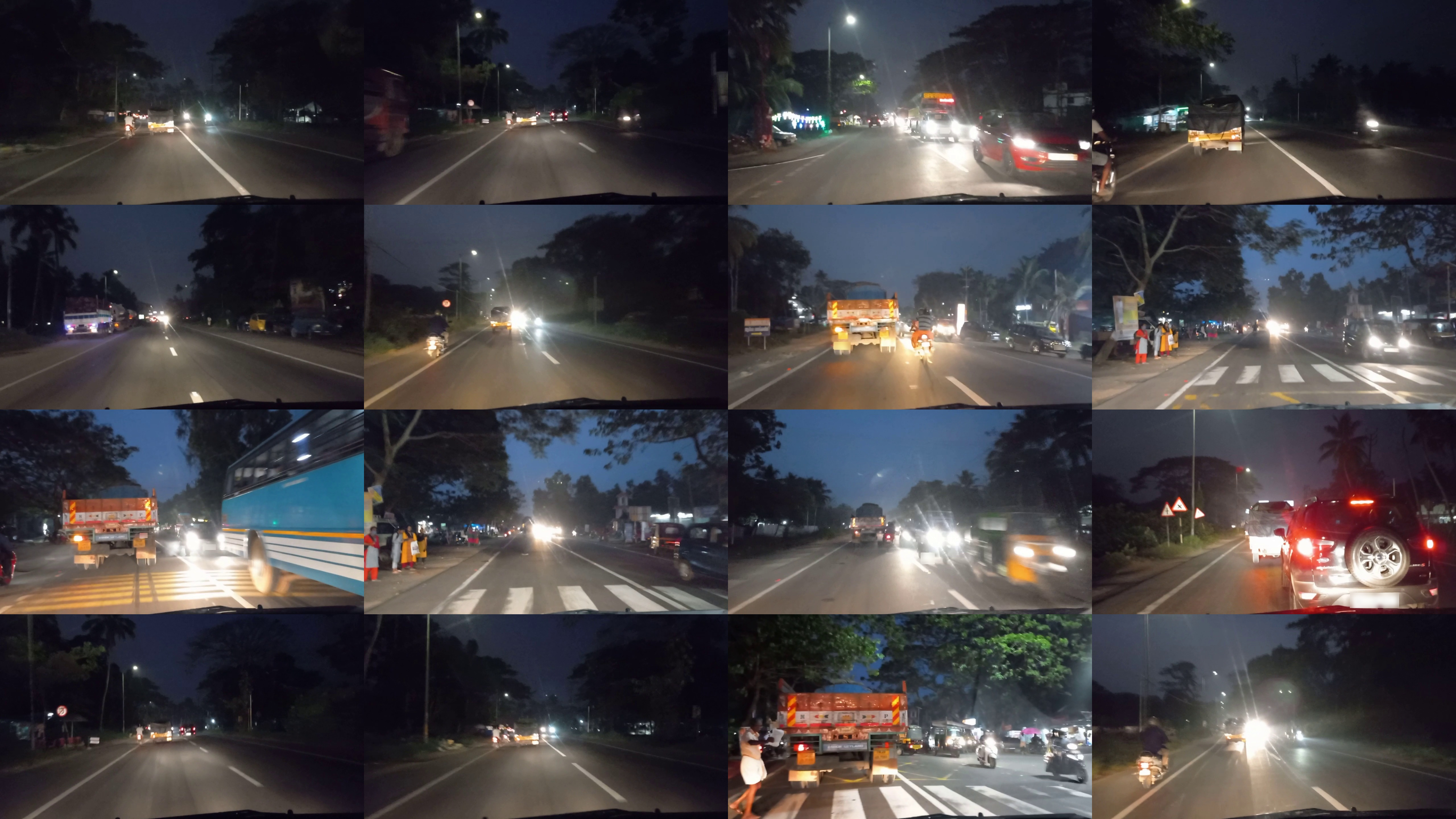}
  \caption {A minimal working example to demonstrate how to place
    two images side-by-side.}
\end{figure*}

\section{Related Work}

\begin{table*}
  \centering
  \begin{tabular}{lllcc}
    \hline
    \textbf{Dataset} & \textbf{Location} & \textbf{\# Images/Scenes} & \textbf{Image QA} \\
    \hline
    Roadscapes & India & 8983 & \checkmark \\
    IDD \cite{idd} & India & 10,004 & x \\
    KITTI \cite{kitti} & Germany & 14,999 & x \\
    rounD \cite{round} & Germany & 13,000+ & \checkmark \\
    BDD-OIA \cite{bddoia} & USA & 11,300 & \checkmark \\
    Lyft Level 5 \cite{lyft} & USA & 55,000+ & \checkmark \\
    RoadQA \cite{roadqa} & China/Asia & 100,000+ & \checkmark \\
    LOKI \cite{loki} & Japan & 1,000+ clips & \checkmark \\
    TITAN-Human Action \cite{titan} & Japan & 1,500+ clips & \checkmark \\
    INTERACTION \cite{interaction} & Global & 11,943 & \checkmark \\
    \hline
  \end{tabular}
  \caption{\label{tab:dataset-comparison}
    Comparative overview of selected driving datasets across multiple tasks and dataset sizes.
  }
\end{table*}

\subsection{Driving Datasets}

A wide range of driving datasets has been developed to support autonomous driving research, each offering distinct features in terms of geographic coverage, scene complexity, and annotation types. The India Driving Dataset (IDD) comprises 10,004 images captured from Indian roads, emphasizing unstructured traffic environments with annotations for object detection, semantic segmentation, and drivable area delineation~\cite{idd}. The KITTI dataset includes 14,999 images from structured European urban and highway scenes and serves as a benchmark for detection and tracking tasks~\cite{kitti}. The rounD dataset provides over 13,000 drone-captured scenes at German roundabouts, offering detailed road user trajectories for behavior analysis~\cite{round}. The Lyft Level 5 dataset contains more than 55,000 scenes from urban environments in the United States, supporting tasks such as detection, tracking, and motion planning~\cite{lyft}. 

\subsection{Driving VQA datasets}

To enhance spatial understanding and reasoning, some datasets incorporate image-based question-answering tasks. The RoadQA dataset includes over 100,000 images and integrates object detection annotations with diverse question-answer pairs, enabling research in spatial reasoning and scene comprehension~\cite{roadqa}. The BDD-OIA dataset, an extension of BDD100K, comprises 11,300 images from U.S. urban environments and introduces object-induced action annotations along with textual explanations, supporting image-level question-answering tasks~\cite{bddoia}.

Video-based question-answering datasets provide temporal context and facilitate higher-level reasoning tasks. The LOKI dataset consists of over 1,000 video clips from urban driving scenes and focuses on intention prediction and language-based reasoning~\cite{loki}. The TITAN-Human Action dataset includes more than 1,500 clips collected in Japanese urban settings, featuring annotations for pedestrian intention, causal reasoning, and video-based question-answering tasks~\cite{titan}.

Visual Question Answering (VQA) has become an increasingly critical area in autonomous driving research. It enhances a model's capacity for both visual perception and reasoning. Recent works have led to several VQA datasets that give us insights into how effective Vision Language models (VLMs) are at understanding and interacting with driving scenes. 

Efforts such as NuScenes-QA is a notable benchmark specifically tailored for multimodal visual question answering in autonomous driving scenarios. Built upon the NuScenes dataset, NuScenes-QA integrates diverse sensor modalities, including camera images and LiDAR point clouds, to enable comprehensive scene understanding. This dataset particularly excels in multimodal reasoning tasks, requiring models to interpret spatial-temporal relationships, object interactions, and environment dynamics across different sensor inputs. Despite these advancements, NuScenes-QA primarily focuses on structured urban environments, limiting its applicability in capturing more diverse or unstructured driving contexts \cite{qian2024nuscenesqamultimodalvisualquestion}.

The DriVQA dataset represents another notable advancement, providing a large-scale benchmark explicitly designed for evaluating driving-specific visual question answering. It includes over 10,000 video sequences annotated with rich question-answer pairs covering diverse reasoning tasks such as action prediction, object localization, and scene understanding. DriVQA uniquely emphasizes temporal and relational reasoning across consecutive frames, thus promoting model capabilities in capturing dynamic visual cues and understanding scene evolution. However, despite its extensive annotations and temporal focus, DriVQA primarily features structured urban driving scenarios and does not fully encompass the complexities and variability found in less structured environments or scenarios involving ambiguous and uncertain driving conditions \cite{drivqa}.

In contrast to prior datasets that predominantly emphasize structured environments and high-end sensor setups, our RoadscapesQA dataset introduces a fresh perspective by capturing the nuanced challenges of unstructured driving conditions found in India. It spans a rich array of urban, rural, and highway settings, including low-light and nighttime scenarios that are frequently underrepresented in existing benchmarks. 

RoadscapesQA goes beyond standard object-level reasoning by incorporating diverse VQA 
tasks such as object counting, spatial relationships, and contextual scene descriptions—all tailored for real-world driving conditions that include unpredictable agent behaviors and varied road infrastructures. This makes RoadscapesQA an important step forward in building VLMs that are not only perceptive but also contextually aware and resilient to the messiness of real-world driving environments.


\section{Roadscapes}
\subsection{Data Collection}
The raw data for the Roadscapes dataset were collected from the cities of Coimbatore and Kochi in India using a monocular action camera, as well as from the national highway connecting them. In total, 5 hours of driving data were recorded, amounting to a total of 35 sequences with an average sequence length of 8 minutes. For data acquisition, a monocular front-facing camera was mounted on the front dash of the vehicle using a camera mount. The data were captured at 30 FPS with a resolution of 1920×1080 pixels. From the raw data, image frames were sampled every 30 frames (17000 images). A number of images in the sequences were unusable because of mild-to-severe distortion. These images were identified and filtered manually from the dataset into 9000 images. The dataset includes annotations for two computer vision tasks: object detection, drivable area segmentation and two multimodal tasks: image-level question answering. It encompasses a diverse range of scenes, including highways, service roads, crowded city streets, and village roads, captured at different times of the day (daytime, dusk, and nighttime). Out of the 35 sequences, 21 sequences are used in the training set and 14 sequences in the validation set. Table 2 shows a detailed breakdown of the dataset statistics based on the scenarios and time of day captured. 

\begin{table*}[ht]
  
  \label{table-dataset-daynight}
  \centering
  \begin{tabular}{l r r}
    \hline
    Day / Night Scenario         & Train & Test \\
    \hline
    Daytime images   & 5,519 & 1,277 \\
    Nighttime images & 1,989 &   196 \\
    Total            & 7,508 & 1,475 \\
    \hline
  \end{tabular}
  \caption{Dataset distribution of images by time of day for Train and Test sets.}
\end{table*}

\subsection{Data Privacy and Anonymization}
In order to maintain the privacy of the subjects within the recorded data, we ran a semi-automated anonymization pipeline to identify the blue 4500 license plates which are personally identifiable and sensitive data. Anonymization was performed using a YOLOv5 detector specialized in license plates~\cite{keremberke_yolov5m_license_plate} and was verified by manual spot checks by sampling 1 in every 100 images in each sequence to ensure compliance. Considering legalities and privacy concerns regarding individuals and vehicles in the dataset, we propose releasing it under an explicit non-commercial license, making it available only to researchers on request.

\subsection{Data Annotation and Generation}
Data annotation performed by humans is typically one of the most resource-intensive aspects of data curation, both in terms of time and cost. Therefore, to reduce the overall time spent by human annotators to label the dataset, we employed the zero-shot YOLOWorld model to capture common object classes like car, truck, bus, motorcycle etc., data annotation using foundation models before the human annotation process. The annotations were verified and improved manually by an annotation team consisting of four individuals from the same academic peer group: three co-authors of this paper and one undergraduate student. The undergraduate annotator was monetarily compensated for the work. For the object detection task, Table 3 depicts the classes present in the dataset. 

\subsection{Visual Question Answering}
In the context of autonomous driving, VQA can help vehicles make informed decisions by answering questions about road conditions, traffic signs, pedestrian behavior, and potential hazards. For the visual question answering dataset, we used rule based heuristics on top of the object detection annotations, to generate ground truth for a variety of questions covering 3 categories: \textbf{Object Counting}, \textbf{Object Description} and \textbf{Surrounding Description}. Each category consists of multiple questions totally adding up to 7 questions per image. Within each category, object classes are selected at random for the generation of the dataset. Table 4 depicts the different types of questions and their generated answer type for the dataset. 

\begin{table}[ht]
  \centering
  
  \label{tab:roadscapes-label-counts}
  \begin{tabular}{l r}
    \hline
    \textbf{Label} & \textbf{Count} \\
    \hline
    motorcycle    & 8,988  \\
    car           & 16,594 \\
    truck         & 11,006 \\
    rider         & 5,675  \\
    person        & 5,847  \\
    traffic sign  & 2,925  \\
    traffic light & 1,322  \\
    bus           & 2,464  \\
    headlight    & 215    \\
    rickshaw      & 1,217  \\
    animal        & 26     \\
    bicycle       & 15     \\
    \hline
  \end{tabular}
  \caption{Roadscapes QA Label Counts}
\end{table}

\begin{table*}
  \centering
  \begin{tabular}{l p{9cm} l}
    \hline
    \textbf{Category} & \textbf{Question} & \textbf{Answer Type} \\
    \hline
    Object Counting & How many objects of type traffic sign are in the image? & Integer \\
    Object Counting & Is there a traffic sign present in the image? & Yes/No \\
    Object Counting & Are there more cars than trucks? & Yes/No \\
    Object Description & What is the color of the truck in the image? & Color \\
    Object Description & What class is the object at bounding box [x, y]? & Object Class \\
    Surrounding Description & What time of day is it? & Time of Day \\
    Surrounding Description & What is the traffic density? & Traffic Level \\
    \hline
  \end{tabular}
   \caption{\label{tab:qa_examples}
    Examples of annotated questions grouped by category and answer type.
  }
\end{table*}

\section{Experimental Setup}

\begin{table*}

  \centering
  \begin{tabular}{lccc}
    \hline
    \textbf{Model} & \textbf{Object Counting Accuracy} & \textbf{Object Description} & \textbf{Surrounding Description} \\
    \hline
    4o         & 0.598  & 0.495 & 0.701 \\
    Paligemma  & 0.187  & 0.501 & 0.485 \\
    Phi-3.5    & 0.667  & 0.437 & 0.643 \\
    4o-mini    & 0.628  & 0.453 & 0.645 \\
    \hline
  \end{tabular}
   \caption{\label{tab:model-performance}
    Performance comparison of four VQA models on the Roadscapes dataset across Object Counting, Object Description (cosine similarity), and Surrounding Description (cosine similarity).
  }
\end{table*}
\subsection{Dataset and Task Categories}
We evaluate vision-language models on the Roadscapes dataset, which comprises three visual question answering (VQA) categories: \textit{Object Counting}, \textit{Object Description}, and \textit{Surrounding Description}. Each category contains 500 questions, providing a diverse set of challenges for autonomous driving scenarios. All models are evaluated in a zero-shot manner, without any fine-tuning on the dataset. The input to each model consists of image-question pairs, and the output is generated as free-form text. This experimental design follows recent VQA benchmarks for autonomous driving, such as LingoQA~\cite{chen2024lingoqa}.

\subsection{Evaluated Models}
We evaluate the following models:
\begin{itemize}
    \item \textbf{Phi-3.5}~\cite{microsoft2024phi3,phi3techreport}: A lightweight, state-of-the-art open multimodal model with strong performance on vision-language reasoning tasks.
    \item \textbf{4o}~\cite{openai2024gpt4o}: A recent multimodal large language model capable of high-quality image and text understanding.
    \item \textbf{Paligemma}~\cite{paligemma2024}: An open multimodal model designed for visual reasoning.
    \item \textbf{4o-mini}~\cite{openai2024gpt4o}: A lightweight multimodal vision-language model variant of GPT-4o, evaluated in a zero-shot setting.

\end{itemize}

\subsection{Evaluation Metrics}
For the Object Counting task, we employ exact-match accuracy as the primary evaluation metric, following established practice~\cite{chen2024lingoqa}. The Object Description and Surrounding Description tasks are evaluated using cosine similarity between sentence embeddings, specifically utilizing the all-MiniLM-L6-v2 model~\cite{wang2020minilm}.

\section{Results}

\subsection{Object Counting}
In the Object Counting task, Phi-3.5 achieved the highest accuracy of 0.667, followed closely by 4o-mini with an accuracy of 0.628. Common failure modes observed across models include undercounting (missing objects in complex scenes), overcounting (double-counting partially occluded objects), and hallucination (reporting non-existent objects).

\subsection{Object Description}
The Object Description task proved more challenging due to the requirement for fine-grained recognition of object attributes. Paligemma demonstrated the best performance with a similarity score of 0.501. Frequent errors included hallucinations of incorrect colors and mislabeling of object classes, highlighting the difficulty of precise object recognition in diverse driving scenarios.

\subsection{Surrounding Description}
This category focused on semantic reasoning tasks, such as determining the time of day or assessing traffic density. The 4o model exhibited the strongest performance, achieving a similarity score of 0.701. Common errors across models included confusion in temporal descriptions (e.g., misinterpreting lighting conditions) and inconsistencies in subjective judgments (e.g., varying assessments of traffic density).

\subsection{Observation}
Our results align with recent findings in autonomous driving VQA~\cite{chen2024lingoqa,parthasarathy2025glimpse}, which report that zero-shot models struggle with fine-grained perception and semantic reasoning in complex scenes. The use of embedding-based metrics for open-ended tasks follows recommendations from prior work~\cite{chen2024lingoqa,wang2020minilm}.

\section{Hallucination Analysis}

\begin{table*}[t]
  \centering
  \begin{tabular}{lccc}
    \hline
    \textbf{Model} & \textbf{Object Counting} & \textbf{Object Description} & \textbf{Surrounding Description} \\
    \hline
    4o         & 21.8\% (109/500) & 51.6\% (258/500) & 7.0\% (35/500) \\
    Paligemma  & 14.6\% (64/439)  & 50.8\% (254/500) & 29.8\% (149/500) \\
    Phi35      & 13.7\% (60/439)  & 52.4\% (262/500) & 6.2\% (31/500) \\
    4o-mini    & 15.4\% (77/500)  & 61.6\% (308/500) & 23.6\% (118/500) \\
    \hline
  \end{tabular}
  \caption{\label{tab:hallucination-rates}
    Hallucination rates (\%) and counts (hallucinations/total) for each model and VQA category.
  }
\end{table*}

\subsection{Overview of Hallucination Detection}

Hallucinations in model outputs were detected using a combination of reference-based and embedding-based methods. For open-ended tasks (Object Description and Surrounding Description), we computed the cosine similarity between sentence embeddings using the all-MiniLM-L6-v2 model~\cite{wang2020minilm}. Predictions with similarity below a calibrated threshold were flagged as hallucinations. For Object Counting, hallucinations were defined as overcounting (predicted count greater than ground truth) or false positives in binary (yes/no) questions. This approach aligns with recent VLM evaluation practices~\cite{chen2024lingoqa,leng2024mitigating}, and is consistent with both reference-based and emerging reference-free hallucination detection frameworks~\cite{li2024referencefree}. The hallucination rate for each category is computed as:

\begin{equation}
\mathrm{Hallucination\ Rate} = \frac{H}{N}
\end{equation}
where $H$ is the number of hallucinated responses and $N$ is the total number of samples.

\subsection{Hallucination Rates Across Models and Tasks}

Table 2 summarizes the hallucination rates (\%) for each model and VQA category. Object Description consistently shows the highest hallucination rates across all models, ranging from 50.8\% to 61.6\%. This suggests that models struggle most with accurately describing specific object attributes, in line with prior findings~\cite{chen2024lingoqa}. Object Counting and Surrounding Description generally exhibit lower hallucination rates, with some exceptions.

Notably, the 4o-mini model demonstrates the highest hallucination rate (61.6\%) for Object Description, significantly higher than other models in this category. Conversely, the 4o and Phi35 models perform relatively well in Surrounding Description tasks, with hallucination rates of 7.0\% and 6.2\%, respectively.

\subsection{Error Patterns and Insights}

Common error patterns observed in hallucinations vary across task categories:
\begin{itemize}
    \item \textbf{Object Counting:} Overcounting and false positives are prevalent, indicating challenges in accurately quantifying objects in complex scenes~\cite{leng2024mitigating}.
    \item \textbf{Object Description:} Hallucinations often involve incorrect colors or misclassified object classes, suggesting difficulties in fine-grained visual perception and attribute recognition~\cite{chen2024lingoqa}.
    \item \textbf{Surrounding Description:} Errors frequently relate to confusion in temporal or contextual reasoning, such as misinterpreting time of day or environmental conditions.
\end{itemize}

These findings highlight the challenges of deploying robust VLMs in real-world autonomous driving scenarios, where accurate perception and reasoning are critical for safety and decision-making~\cite{zhai2023survey}.

\subsection{Comparison to Related Work}

The observed hallucination rates align with recent studies on VLM evaluation in autonomous driving contexts. For instance, the high hallucination rates in Object Description tasks (50.8\%--61.6\%) are consistent with challenges reported in fine-grained attribute recognition tasks in LingoQA and VL-CheckList evaluations~\cite{chen2024lingoqa,li2024referencefree}. However, the relatively low hallucination rates in Surrounding Description tasks for some models (e.g., 4o at 7.0\% and Phi35 at 6.2\%) suggest potential improvements in contextual reasoning compared to previous benchmarks. This progress may be attributed to advancements in model architectures and training techniques~\cite{leng2024mitigating}.

These results underscore the importance of targeted evaluation strategies for VLMs in autonomous driving applications. Future model development should focus on reducing hallucinations in object attribute recognition and improving consistency in contextual reasoning to enhance the reliability and safety of VQA systems in real-world driving scenarios.


\section{Conclusion}
This research addresses a critical gap in large-scale image question answering (IQA) datasets for driving scenes in southern India. The Roadscapes dataset fills this void by providing comprehensive VQA data for the southern part of India, including the previously underrepresented Coimbatore–Kochi corridor. This expansion complements existing datasets like IDD and enables new types of vision-language evaluation.

Our analysis of hallucination rates across multiple models using embedding-based and counting metrics has revealed key challenges in model reliability for Indian driving scenarios. The findings highlight varying performance across different task types, with Object Description tasks presenting the highest hallucination rates. These insights underscore the complexities of deploying vision-language models in real-world autonomous driving contexts.

Roadscapes' unique contribution lies in its coverage of the Coimbatore–Kochi region and the inclusion of VQA tasks not available in existing road scene datasets. This comprehensive approach enables more robust benchmarking of vision-language models for Indian roads, supporting the development of safer and more context-aware models for underrepresented regions. 

\section{Limitations}

While the Roadscapes datasetrepresents a significant advancement in image question answering for driving scenes in southern India, several limitations should be acknowledged:

\begin{enumerate}
    \item \textbf{Limited task coverage:} The current dataset does not include explicit tasks or benchmarks for object localization or spatial relations, despite their importance in autonomous driving scenarios~\cite{idd,round}. Future work could address this by incorporating these tasks into the dataset and evaluation framework. The dataset lacks specific benchmarks for distinguishing spatial relationships (e.g., ``left'' from ``right''). Future iterations could leverage powerful embedding models with spatial language understanding capabilities to address this limitation~\cite{leng2024mitigating,li2024referencefree,zhai2023survey}.

    \item \textbf{Geographic coverage:} Although the dataset covers the 
    Coimbatore-Kochi corridor, broader geographic coverage is needed to ensure even greater diversity and generalizability across all regions of India. Expanding the dataset to include more diverse driving environments would enhance its representativeness~\cite{idd,arxivsurvey}.
    
    \item \textbf{Annotation types and task complexity:} The dataset could benefit from more complex vision-language tasks to further enhance its utility for the research community. This could include multi-turn dialogues, temporal reasoning, or multi-image tasks~\cite{chen2024lingoqa}.
\end{enumerate}

Addressing these limitations in future work will strengthen the Roadscapes dataset's contribution to autonomous driving research and vision-language model development for diverse global contexts.



\bibliography{custom}

\appendix




\end{document}